RESEARCH ARTICLE                                           OPEN ACCESS

# Comparative Analysis of Deep-Fake Algorithms

Nikhil Sontakke [1], Sejal Utekar [2], Shivansh Rastogi [3], Shriraj Sonawane [4]

[1] Department of Computer Engineering, Vishwakarma Institute of Technology, Pune Maharashtra
[2] Department of Computer Engineering, Vishwakarma Institute of Technology, Pune Maharashtra
[3] Department of Computer Engineering, Vishwakarma Institute of Technology, Pune Maharashtra
[4] Department of Computer Engineering, Vishwakarma Institute of Technology, Pune Maharashtra

**ABSTRACT**

Due to the widespread use of smartphones with high-quality digital cameras and easy access to a wide range of software apps for recording, editing, and sharing videos and images, as well as the deep learning AI platforms, a new phenomenon of 'faking' videos has emerged. Deepfake algorithms can create fake images and videos that are virtually indistinguishable from authentic ones. Therefore, technologies that can detect and assess the integrity of digital visual media are crucial. Deepfakes, also known as deep learning-based fake videos, have become a major concern in recent years due to their ability to manipulate and alter images and videos in a way that is virtually indistinguishable from the original. These deepfake videos can be used for malicious purposes such as spreading misinformation, impersonating individuals, and creating fake news. Deepfake detection technologies use various approaches such as facial recognition, motion analysis, and audio-visual synchronization to identify and flag fake videos. However, the rapid advancement of deepfake technologies has made it increasingly difficult to detect these videos with high accuracy. In this paper, we aim to provide a comprehensive review of the current state of deepfake creation and detection technologies. We examine the various deep learning-based approaches used for creating deepfakes, as well as the techniques used for detecting them. Additionally, we analyze the limitations and challenges of current deepfake detection methods and discuss future research directions in this field. Overall, the paper highlights the importance of continued research and development in deepfake detection technologies in order to combat the negative impact of deepfakes on society and ensure the integrity of digital visual media.

## I. INTRODUCTION

Deepfake detection is a relatively new field of research that emerged in response to the growing threat of manipulated media, particularly videos, in which the appearance and/or behavior of individuals is artificially altered using deep learning techniques. The term "deepfake " was first coined in 2017, and since then, the technology has evolved rapidly, making it easier and cheaper to create convincing deepfake videos.

Technological advancements, particularly in handheld devices with high-definition cameras, combined with the widespread use of artificial intelligence tools, models, and apps, have resulted in a large number of videos of world-famous celebrities and leaders that have been doctored to convey fake news for political gain or to ridicule specific individuals [1]. Photographic and video evidence are routinely utilized in courtrooms and police investigations and are considered to be trustworthy. Video evidence, however, is becoming potentially untrustworthy as video manipulation techniques progress. It's likely that, in the not-too-distant future, video evidence will need to be reviewed for signs of tampering before being considered acceptable in court.

To train models to create photorealistic images and videos, deepfake methods often require a vast amount of image and video data. Celebrities and politicians are the first targets of deepfakes since they have a vast amount of videos and photographs online. In pornographic photographs and movies, deepfakes were utilized to replace the faces of celebrities and politicians with bodies. In 2017, the first deepfake video was released, in which a celebrity's face was replaced with a porn actor's face. Deepfake methods can be used to make movies of world leaders with fake speeches for the aim of falsification, which poses a threat to global security. In 2018, a fake video was made of Barack Obama, using statements he never said. Furthermore, DeepFakes have already been used to distort Joe Biden footage showing his tongue out during the US 2020 election. These detrimental applications of deepfakes can have a significant impact on our society and can lead to the spread of false information, particularly on social media [3].

Several big companies have decided to take action against this phenomenon, Google has created a database of fake videos to support researchers who are developing new techniques to detect them, while AWS, Facebook, Microsoft, the Partnership on AI's Media Integrity Steering Committee, and academics have come together to build the Deepfake Detection Challenge (DFDC). The goal of the challenge is to spur researchers around the world to build innovative new technologies that can help detect deep fakes and manipulated media.

Deepfake detection is a rapidly growing field of research that aims to identify and mitigate the threat of manipulated media, particularly videos, in which the appearance and/or behavior of individuals is artificially altered using deep learning techniques. The use of deepfake technology has the potential to cause significant harm in areas such as politics, entertainment, and personal relationships. In response, researchers have developed various algorithms and techniques using artificial intelligence (AI) to detect deepfakes and preserve the integrity of digital media. This paper aims to provide an overview of the current state-of-the-art in deepfake detection, including the most widely used algorithms and approaches in the field. The goal of this paper is to provide a comprehensive introduction to deepfake detection using AI and to highlight the challenges and opportunities in this rapidly evolving field of research.

We examined various deepfake strategies for the development and detection of deepfakes in our article.

## II. LITERATURE REVIEW

Thanh Thi Nguyen, Cuong M. Nguyen, Dung Tien Nguyen, Duc Thanh Nguyen and Saeid Nahavandi mentioned that Deep learning has been used to handle a variety of complicated challenges, including large data analytics, computer vision, and human-level control. Deep learning technologies, on the other hand, have been used to develop software that poses a risk to privacy, democracy, and national security. "deepfake" is a recent example of a deep learning-powered application. Deepfake algorithms can make fake photos and videos that people can't tell apart from the real thing. As a result, the development of tools that can detect and analyze the integrity of digital visual material is critical. Their study included a survey of deepfake-creation algorithms and, more crucially, deepfake detection methods proposed in the literature to date. They hold in-depth conversations on the difficulties, research trends, and future prospects of deepfake technologies [1].

Koopman, Marissa, Andrea Macarulla Rodriguez, and Zeno Geradts in their paper stated that deep fake poses a threat to global security and integrity of multimedia, the Deepfake algorithm allows a user to photorealistically swap the face of one actor in a video with the face of another. This raises forensic issues in terms of video evidence dependability. Photo response non uniformity (PRNU) analysis is examined for its effectiveness at identifying Deepfake video tampering as part of a solution. The PRNU study reveals a substantial difference between authentic and Deepfake videos in mean normalized cross correlation scores [4].

Lyu, Siwei stated that the status of films and audios as definitive evidence of events has begun to be challenged by high-quality fake videos and audios generated by AI algorithms (deep fakes). They explored research opportunities in this area and highlighted a number of these problems in their publication [5].

Guarnera, Luca, Oliver Giudice, and Sebastiano Battiato in their paper mentioned deep learning is a powerful and versatile technology that has been widely used in domains such as computer vision, machine vision, and natural language processing. Deepfakes employs deep learning technology to modify photographs and videos of people to the point where humans are unable to distinguish them from the genuine thing. Many studies have been undertaken in recent years to better understand how deepfake function, and many deep learning-based algorithms have been presented to detect deepfake videos or photos. They conducted a thorough review of deepfake production and detection technologies utilizing deep learning methodologies in this paper. They also provided a thorough examination of various technologies and their use in the identification of deepfake [3].

The Deepfake phenomena has exploded in popularity in recent years as a result of the ability to make very realistic images using deep learning methods, mostly adhoc Generative Adversarial Networks (GAN). They concentrate their research on the study of Deepfakes of human faces with the goal of developing a novel detection approach that can detect a forensics trail buried in photos, similar to a fingerprint left in the image generating process. They introduced a technique that extracts a set of local characteristics specially targeted to represent the underlying neural generation process using an Expectation Maximization (EM) algorithm. Experimental experiments with naive classifiers on five alternative architectures (GDWCT, STARGAN, ATTGAN, STYLEGAN, STYLEGAN2) on the CELEBA dataset as ground-truth for non-fakes were used for ad-hoc validation [6].

## III. METHODOLOGIES

Deepfake is a methodology for creating false photos and videos that employs the concepts of Generative Adversarial Networks (GANs). We'll start with an overview of the current applications and technologies for creating deepfake images and videos in this part. Then, to address this issue, we describe different deep learning detection strategies.

### A. DeepFake Generation Methodologies:

Deep fakes can be generated and detected using a variety of models. GAN (Generative Adversarial Networks) is a model that generates Deep Fakes . GANs, or Generative Adversarial Networks, is a type of generative modeling that employs deep learning techniques such as convolutional neural networks.

Generative modeling belongs to the unsupervised learning category in machine learning this model can then be used to discover and learn regularities or patterns in input data so that it can be applied to the input data of the system and the model may be used to produce or output new examples that could have been drawn from the original dataset.

As generative models, GANs are exciting and rapidly advancing, promising to provide realistic examples across a wide range of problem domains, most notably in image-to-image translation tasks, such as converting images of winter or summer to day and night, and in generating photorealistic photos of objects, scenes, and people that even humans cannot discern as fake.

supervised vs. unsupervised learning and discriminative vs. generative modeling are discussed in the context of GANs.

GANs are a system for automatically training a generative model by treating an unsupervised problem as supervised and employing both a generative and a discriminative model.

GANs offer a route to advanced domain-specific data augmentation as well as a solution to problems that demand generative solutions, such as image-to-image translation.

The GAN model architecture is made up of two sub-models i.e a generator model for creating new instances and a discriminator model for determining whether the generated examples are fake, generated by the generator model, or real generated from the domain. Generator is the model used to produce fresh credible instances from the area of the problem and Discriminator Model is used for determining if instances are genuine (from the domain) or not (generated).

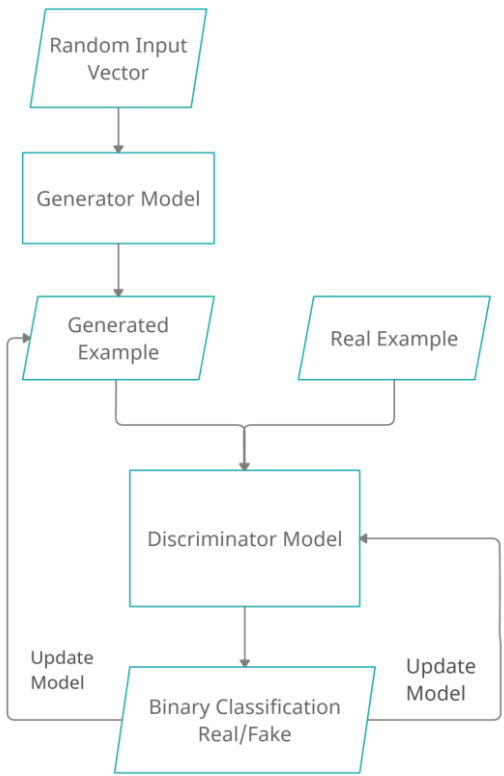

Fig. 1. Working of Generative Adversarial Network Model

The Generator Model:

An input random vector of fixed length is used as input for the generator model in order to generate a sample in the domain.

A Gaussian distribution is employed to generate the vector, which is then used to seed the generative process. Points in this multidimensional vector space will correspond to points in the issue domain during training, resulting in a compressed representation of the data distribution.

This vector space is called a latent space (a latent variable is a random variable that we can't see clearly) or a vector space made up of latent variables. Latent variables, often known as hidden variables, are variables that are relevant to a domain but are not easily visible.

A latent space is a compressed or high-level idea of observable raw data, such as the distribution of input data. In the case of GANs, the generator model assigns meaning to points in a given latent space, allowing new points to be drawn from the latent space as input and utilized to produce new and varied output examples.

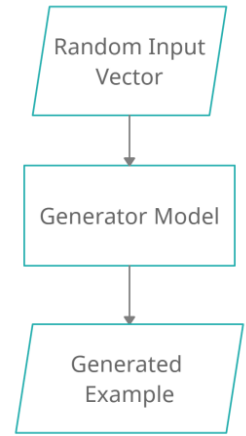

Fig 2. Generator Model of GAN

The Discriminator Model:

The discriminator model takes a domain example (actual or produced) as input and predicts whether it is real or fake (generated).

The real-world example is taken from the training data set. The generator model produces the generated examples. A conventional (and well-understood) classification model serves as the discriminator. The discriminator model is destroyed after the training procedure because we are only interested in the generator.

Because it has learned to extract characteristics from examples in the issue area, the generator can sometimes be repurposed. Using the same or similar input data, some or all of the feature extraction layers can be employed in transfer learning applications.

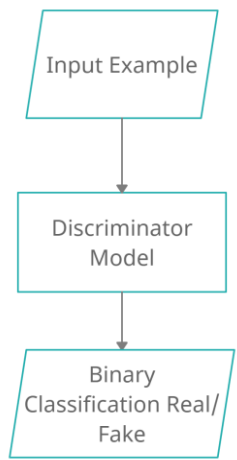

Fig. 3. Discriminator Model of GAN

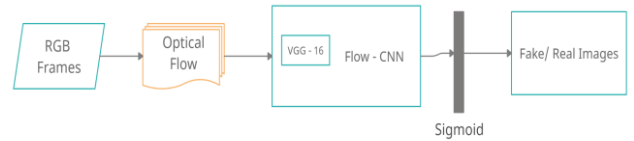

ig. 4. Working of CNN model

*B. DeepFake Detection Techniques*

DeepFakes employ an unique approach that involves modifying fixed areas on the face that must be used as a foundation for superimposition. The algorithm generates different deepfakes in a similar fashion, resulting in some differences during the editing process.

Compression differences, illumination differences, and temporal discrepancies such as lip and eye movements can all be used to train algorithms to detect DeepFake movies [14]. Below mentioned are some techniques that are used for DeepFake detection.

*1) Convolution Neural Network :-*

Convolution Neural Networks (CNN) have been a popular choice among the approaches offered for DeepFake detection. When compared to other approaches for supervised learning in Artificial Intelligence, CNNs have shown significant aptitude and scalability for applications involving image and video operations. CNN has the unique ability to extract information from images that can subsequently be used in a variety of ways. Other supervised learning technologies can then be utilized for final classification for DeepFake to produce stronger and more precise models for DeepFake Detection.

CNNs have an input and output layer, as well as one or more hidden layers, similar to neural networks. The inputs from the first layer are read by the hidden layers, which then apply a convolution mathematical process on the input values. Convolution denotes a matrix multiplication or other dot product in this context. Following matrix multiplication, CNN employs a nonlinearity activation function like the Rectified Linear Unit (RELU), followed by additional convolutions such as pooling layers. Pooling layers' main purpose is to minimize the dimensionality of data by computing outputs using functions like maximum pooling or average pooling.

Transfer learning can be used to detect DeepFakes for practical reasons. Transfer learning makes use of pre-trained neural network weights to train a fine-tuned version of the same on a different dataset for a given purpose. VGG Net, Xception, Inception, and ResNet are just a few of the pre-trained models that have been given open source. A fine-tuned model based on a pre-trained convolutional network, such as the VGG Net, is offered for Image Analysis on human faces.

*2) RNN (Recurrent Neural Networks):-*

Long Short Term Memory (LSTM) networks are a form of Recurrent Neural Network (RNN) that were first used to learn long-term data dependencies. When a deep learning architecture includes both an LSTM and a CNN, it is referred to as "deep in space" and "deep in time," which are two distinct system modalities. In visual identification tasks, CNNs have had a lot of success, while LSTMs are commonly utilized for long sequence processing difficulties. A convolutional LSTM architecture has been extensively studied for other computer vision tasks involving sequences (e.g. activity recognition or human re-identification in videos) and has resulted in significant improvements due to its inherent properties (rich visual description, long-term temporal memory, and end-to-end training).

Another application of artificial neural networks is the recurrent neural network (RNN), which can learn characteristics from sequence data. RNN is made up of numerous invisible layers, each with a weight and bias, similar to neural networks. Relationships between nodes in a direct cycle graph that run in order in RNN. RRN has the advantage of allowing temporal dynamic behavior to be discovered. RNNs, unlike feed forward networks (FFNs), use an internal memory to remember sequences of information from earlier inputs, making them helpful in a range of applications, such as natural language processing and audio recognition. A temporal sequence can be handled by an RNN by introducing a recurrent hidden state that captures interdependence across time scales.

*3) LSTM( Long short term memory)*

LSTM is a type of artificial recurrent neural network (RNN) that handles long-term dependencies. The LSTM uses feedback connections to learn the whole data stream. LSTM has been used in a variety of domains that use time series data, such as classification, processing, and prediction. LSTMs have a common architecture that includes an input gate, a forget gate, and an output gate. The cell state is a type of long-term memory that remembers and saves values from prior intervals in the LSTM cell. The input gate, for starters, is in charge of picking the values that should be entered into the cell state. By using a sigmoid function with a range of, the forget gate is capable of deciding which information to forget. The output gate specifies which current-time information should be taken into account in the next phase.

A convolutional LSTM is used to provide a temporal sequence descriptor for image manipulation of the shot frame given an image sequence. An integration of fully-connected layers is employed to convert the high-dimensional LSTM description to a final detection probability, aiming for end-to-end learning. To reduce training over-fitting, our shallow network comprises two fully connected layers and one dropout layer. A CNN and an LSTM are two types of convolutional LSTM.

For Sequence Processing, using LSTM. Assume a 2-node neural network with the probability of the sequence being part of a DeepFake video or an untampered video as input and a series of CNN feature vectors of input frames as output. The main problem we must solve is the creation of a model that can recursively analyze a sequence in a meaningful way. We employ a 2048-wide LSTM unit with a 0.5 risk of dropout to solve this challenge, which is capable of doing exactly what we require. Our LSTM model, in particular, takes a sequence of 2048-dimensional ImageNet feature vectors during training. A 512 fully-connected layer with a 0.5 risk of dropout follows the LSTM.

Finally, we compute the odds of the frame sequence being pristine or deepfake using a softmax layer. The LSTM module is an intermediary unit in our pipeline that is trained from start to finish without the need of auxiliary loss functions.

## IV. DATASETS FOR DEEPFAKE DETECTION

There are seven publicly available datasets for deepfake detection which can be very helpful for researchers - FFHQ, 100K-Faces, DFFD, CASIA-WebFace, VGGFace2, The eye-blinking dataset, DeepfakeTIMIT.

### A. Flickr-Faces-HQ, FFHQ

Karras et al.[7] proposed a human face dataset (Flickr-Faces-HQ, FFHQ). The FFHQ dataset contains a collection of 70,000 high-resolution face photos generated using generative adversarial networks (GAN). The photographs were gathered via the Flicker platform and include images of people wearing eyeglasses, sunglasses, hats, and other accessories. According to the author, the dataset was pre-processed to reduce the collection and eliminate noise from the photos.

### B. 100K-Faces

100K-Faces [8] is a well-known public dataset that contains 100,000 unique human photos created with StyleGAN [7]. StyleGAN was used to create shots with a flat background from a big dataset of over 29,000 images obtained from 69 different models.

### C. Fake Face Dataset (DFFD)

Dang et al. [9] just published a new dataset named Diverse Fake Face Dataset (DFFD). DFFD contains 100,000 and 200,000 fake images created using state-of-the-art technologies, respectively (ProGAN and StyleGAN models). The collection contains roughly 47.7% male photographs and 52.3 percent female photographs, with the majority of the samples ranging in age from 21 to 50 years old.

### D. CASIA-WebFace

Dong et al. [10] published the CASIA-WebFace database, which has approximately 10,000 people and 500,000 photographs. This information was gathered from the IMDB database, which covers 10,575 well-known actors and actresses. After that, clustering methods are used to retrieve the images of such celebs.

### E. VGGFace2

Cao et al. [11] introduced the VGGFace2 dataset, which is a large-scale face dataset. Over three million face photographs from over nine thousand different persons are included in this collection, with an average of more than 300 images per subject. Images were acquired using the Google search engine, which contains a wealth of information such as ethnicity, lighting, age, and occupation (e.g., actors, athletes, and politicians).

### F. The Eye-Blinking Dataset

The currently available dataset was not created with eye-blinking detection in mind. Li et al. [13] published eye-blinking datasets that were created specifically for this purpose. This dataset contains 50 interviews and films for each participant, each lasting approximately thirty seconds and involving at least one eye blinking. The author then tags the left and right eye states for each video clip using their own tools.

### G. DeepfakeTIMIT

Korshunov et al. created a dataset of videos called DeepfakeTIMIT. [12] uses the database to create a collection of swapped face movies using the GAN-based method. A lower quality model with 64 64 input/output size and a higher quality model with 128 128 input/output size were used to create the dataset. There are 32 subjects in each non-real video collection. For each subject, the author made ten fake videos.

## V. COMPARATIVE ANALYSIS OF DEEPFAKE DETECTION

Following is the thorough analysis of deepfakes creation & detection approaches on the basis of parameters like Architecture, Network structure, Input/Output, Variants, and their use cases.

TABLE I
ANALYSIS OF MAJOR DEEPFAKE ALGORITHMS

| Parameters | GAN | CNN | RNN |
|---|---|---|---|
| Architecture | Deep learning algorithmic architecture that uses two neural networks. | Feed-forward neural networks using filters and pooling | Recurring network that feeds the results back into network |
| Network Structure | Discrimination model, Generation model | Input Layer, Convolution Layer, Pooling Layer, Full connected layer | Input layer, Hidden Layer, output layer |
| Input/Output | The size of input is fixed | The size of the input and resulting output are fixed | The size of the input and the resulting output may vary |
| Variants | DCGAN | LeNet, AlexNet, VggNet | LSTM |
| Use Cases | Image generation, Video generation | Image recognition and classification, face detection, medical analysis, drug discovery | Text translation, natural language processing, conversational intelligence, sentiment analysis. |

TABLE II
ANALYSIS OF VARIANTS OF DEEPFAKE DETECTION ALGORITHMS

| CNN Variant | Accuracy | Precision | F1-score | AUC |
|---|---|---|---|---|
| VGG19 | 0.94 | 0.91 | 0.94 | 0.987 |
| VGG16 | 0.92 | 0.93 | 0.92 | 0.977 |
| VGGFace | 0.99 | 0.99 | 0.99 | 0.998 |
| DenseNet169 | 0.95 | 0.99 | 0.95 | 0.996 |
| DenseNet201 | 0.96 | 0.96 | 0.96 | 0.994 |
| DenseNet121 | 0.97 | 0.99 | 0.82 | 0.971 |
| ResNet50 | 0.97 | 0.99 | 0.97 | 0.997 |
| LSTM | 0.94 | 0.85 | 0.98 | 0.893 |

## VI. CONCLUSIONS

Deepfake is a new technique for deceiving a huge number of people. Though not all deepfake contents are harmful, they must be identified because some of them are actually dangerous to the globe. The major goal of this research was to analyze different methods available for detecting deepfake photos and videos. Using a variety of approaches, several other academics have been working tirelessly to detect deepfake content. Their ground-breaking work will have a huge impact on our society. Deepfake victims can rapidly evaluate whether the images are real or fake with this technology. People will be attentive since they will be able to recognise a deepfake image thanks to our efforts.

Many more trials and tests will be carried out in the future and as advancements occur in this technology we may be able to use more efficient models to detect deepfake photos and videos in order to reduce crime in our community and, more broadly, the world and strengthen global security.